\documentclass{article}
\usepackage{spconf,amsmath,graphicx}
\usepackage{multirow}
\usepackage{todonotes}
\usepackage{pifont}
\newcommand{\xmark}{\ding{55}}%

\title{End-to-End Speech Summarization using Restricted Self-Attention}
\name{Roshan Sharma, Shruti Palaskar, Alan W Black and Florian Metze }
\address{
	Carnegie Mellon University\\
	Pittsburgh PA,USA\\
	\texttt{roshansh@andrew.cmu.edu}, \texttt{spalaska@andrew.cmu.edu}}

\begin{document}
%
\maketitle
\message{\showthe\columnwidth}
\begin{abstract}
Speech summarization is typically performed by using a cascade of speech recognition and text summarization models. End-to-end modeling of speech summarization models is challenging due to memory and compute constraints arising from long input audio sequences. Recent work in document summarization has inspired methods to reduce the complexity of self-attentions, which enables transformer models to handle long sequences.
In this work, we introduce a single model optimized end-to-end for speech summarization. We apply the restricted self-atention technique from text-based models to speech models to address the memory and compute constraints. We demonstrate that the proposed model learns to directly summarize speech for the How-2 corpus of instructional videos. The proposed end-to-end model outperforms the previously proposed cascaded model by 3 points absolute on ROUGE. Further we consider the spoken language understanding task of predicting concepts from speech inputs, and show that the proposed end-to-end model outperforms the cascade model by 4 points absolute F-1.
\end{abstract}
\begin{keywords}
speech summarization, end-to-end , long sequence modeling, concept learning
\end{keywords}

\section{Introduction}
\label{sec:intro}
Summarization extracts and condenses desired information from the inputs, often text. Text can be summarized using abstraction or extraction \cite{Hahn2002}. Abstractive Text Summarization (ATS), generates a novel and concise summary of the input text. Abstractive summarization can be performed on multiple modalities \cite{palaskar-etal-2019-multimodal,yu2021vision}.

Speech Summarization is performed using a cascade of Automatic Speech Recognition (ASR) followed by Abstractive Text Summarization (ATS) \cite{zhu2020hierarchical,rezazadegan2020automatic,manakul2020abstractive}. \cite{palaskar21_interspeech} proposed an alternative cascade formulation- ASR followed by Concept Extraction and Summarization.They showed that specific and abstract concepts are useful as intermediate representations for multimodal summarization. However, cascade architectures makes the model structure complicated, and errors in the ASR degrade summarization performance. Therefore, we propose a single sequence model optimized end-to-end (E2E) for speech summarization.

Speech summarization involves very long input sequences. The prohibitive quadratic computational cost of self-attention makes standard transformer models unsuitable for longer sequences. To address this,   \cite{dai-etal-2019-transformer} uses segment-wise recurrence within transformer self-attention to provide longer context, and \cite{Rae2019CompressiveTransformer} compresses the segment level contexts and provides them as additional input to enable a longer context. Other works have focused on making the self-attention sparse to reduce computational complexity. . Reformer \cite{kitaev2019reformer} uses Locality Sensitive Hashing to compute localized self-attention in O(n.logn) and ETC \cite{ainslie-etal-2020-etc} uses efficient global-local attention to scale to longer sequences. To reduce the complexity of self-attention to O(n), Linformer \cite{wang2020linformer} uses a low-rank factorization of the self-attention matrix, and Big Bird \cite{Zaheer2020BigBird} uses a combination of sliding window, global and random attention. Longformer  \cite{Beltagy2020Longformer} uses different attention patterns for each layer and restricted dilated self-attention with task-specific global attention. These long sequence techniques have been evaluated on text inputs, where the input sequence lengths are often several hundred times smaller than sequence lengths of video-level speech (see Table \ref{tab:how2_datastats}).

Abstractive speech summarization uses the complete long sequence context to generate a summary. Long context ASR has been explored by training on longer sequences \cite{Shafey2019,HoriMHR20,hori2021advanced} or by passing context across utterances \cite{Masumura2021}. \cite{Shafey2019} trains contextual models for joint ASR-diarization by concatenating turns from doctor-patient conversations. \cite{HoriMHR20,hori2021advanced} trains long context transformer ASR models by concatenating text and audio from previous utterances as input to decode the current utterance. Longer segments produce larger WER improvements and using both acoustic and lexical context is shown to be important. In this work, we 
 \begin{enumerate}
     \item introduce a way to directly model speech summarization as an end-to-end task
     \item  demonstrate the effectiveness of restricted self-attention for speech inputs, critical for the success of end-to-end speech summarization, and 
     \item show that such an end-to-end model can also be applied to learn concepts directly from speech inputs, a potential spoken language understanding task. 
 \end{enumerate}


\section{Background}
\label{sec:background}

\subsection{Cascade and E2E Modeling}

\begin{figure}
    \centering
    \includegraphics[width=0.5\textwidth]{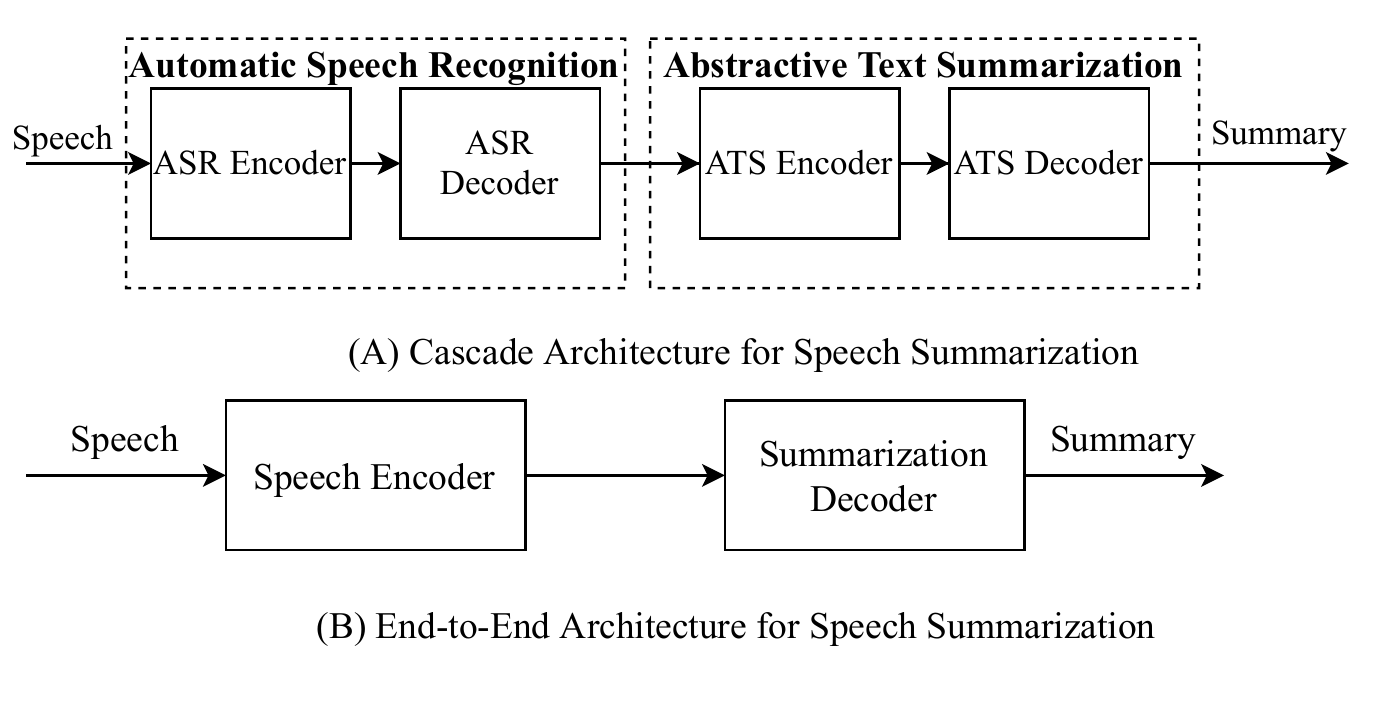}
    \caption{Speech Summarization: Cascade and End-to-End Model Architectures}
    \label{fig:architectures}
\end{figure}

Speech summarization can be modeled as a cascade of speech recognition and text summarization \cite{zhu2020hierarchical,rezazadegan2020automatic}. Figure \ref{fig:architectures} shows the cascade and end-to-end approaches to speech summarization. The cascade approach benefits from strong ASR models pre-trained on large amounts of speech and summarization models like BART \cite{lewis-etal-2020-bart} trained on large amounts of text data. However, errors in ASR are compounded due to the cascade architecture, which serves as motivation for direct end-to-end modeling.

\subsection{Concept Learning}


Cascaded learning was recently shown as one of the ways to accomplish speech summarization \cite{palaskar21_interspeech}. In \cite{palaskar21_interspeech}, authors propose multimodal speech summarization via semantic concept learning, where speech is first represented as a sequence of concepts, followed by a concept to summarization model that generates abstractive summaries. Concepts are abstractive representations of the input speech (and video) that can be used as anchoring points between speech and the summary. Domain-specific noun phrases are automatically extracted and used as abstract concepts (extracted from the human-annotated summaries for the training data). Their model is a pipeline model that does not attempt end-to-end generation of summaries given input speech. In this work, we attempt to model speech summarization as an end-to-end task and also evaluate the benefits of concept learning for such end-to-end speech summarization.

\begin{table}[h]
\caption{Statistics of the How-2 2000h Dataset used for model training and evaluation. The mean and maximum statistics of N- the input length in frames, and L- the output length (in tokens) is shown.}
\resizebox{0.4\textwidth}{!}{%
\centering
\begin{tabular}{lcccc}
\hline
\textbf{Set} & \textbf{Max N} & \textbf{Mean N} & \textbf{Mean L} & \textbf{Max L} \\ \hline
Train          &      145,082          &    9,806.58                                                                &  60.54                    &           173          \\ \hline
Test               &     39,537         &  9,866.55                                                                  &    60.29                  &   152                  \\ \hline
\end{tabular}
}
\label{tab:how2_datastats}
\end{table}




\section{Proposed Approach}
\label{sec:proposed}
\subsection{Restricted Self-Attentions}
Different from other speech tasks like Automatic Speech Recognition or Speech Synthesis, the speech inputs for summarization are much longer. Table \ref{tab:how2_datastats} shows the average and maximum frame lengths of input speech, and output token lengths for summarization. The input speech segments measure over 100s, whereas typical ASR utterance lengths are around 5s. 
The high computational complexity makes it intractable to train video-level speech models on a 32 GB GPU. Consider $N$ is the length of the input speech sequence, and $L$ is the length of the output token sequence( $N$ $>>$ $L$, refer Table \ref{tab:how2_datastats} for values) . Then, there are three attentions in the sequence model: encoder self-attention with a computational complexity of $O(N^2)$, decoder self-attention with a complexity of $O(L^2)$, and encoder-decoder source-target attention with a complexity of $O(NL)$. In order to make end-to-end training possible, the computational complexity of the encoder self-attention needs to be reduced. 

Inspired by \cite{Beltagy2020Longformer,Moritz2021}, we break down the self-attention computation into fixed sized context windows of size $W$. For each sequence element, a surrounding context of width $W/2$ on each side is considered while computing the self-attention result. The number of such windows required will be $P=N/W$, and the cost of the encoder-self attention is now reduced to $O(PW^2)$, which is smaller than $O(N^2)$. To further reduce the computational complexity, we can drop one element for every $D$ elements, i.e., dilation. Dilation further reduces the complexity to $O(P(W/D)^2)$.

\subsection{End-to-End Speech Summarization}

Given input speech frames for an \textit{entire} video, we propose to directly summarize it into short, abstractive, textual summaries. The objective of mapping long speech frames (details in Table \ref{tab:how2_datastats}) onto significantly shorter textual tokens makes this an End-to-End Speech Summarization task. As training summarization models from scratch is challenging, we pretrain the sequence model using ASR. Then, the encoder-decoder model is fine-tuned for speech summarization. 



\subsection{End-to-End Concept Learning}


Semantic concepts were shown to be a strong grounding aspect across modalities, especially to bridge the gaps in cascaded speech summarization \cite{palaskar21_interspeech}. Intermediate concept learning can be useful for controllability of generated summaries. Abstract concepts were extracted in \cite{palaskar-etal-2019-multimodal} by transcribing the videos into text format, and then training a concept extractor. We contend that it would be useful to train a concept extractor from speech end-to-end. As we propose end-to-end speech summarization, we also evaluate the usability of our model to generate concepts directly from speech. 

Given input speech at the video-level, we generate abstract semantic concepts as outputs. Sequence of abstract concepts is generated as natural language text. During inference, this model generates such semantic concepts directly from speech, which can be used further for downstream summarization in a cascaded setup.



\section{Experimental Setup}
\label{sec:experiments}
\subsection{Dataset and Evaluation}
The How-2 Dataset \cite{sanabria2018how2} contains 2000h of instructional videos with corresponding text transcripts, video, speech, translations, and summaries. 

Two tasks are evaluated: (a) Abstract Concept Generation from Speech, and (b) Abstractive Speech Summarization. Concept Generation is evaluated using Precision, Recall, and F-1 score. Summarization is evaluated using ROUGE \cite{lin-2004-rouge}, METEOR \cite{denkowski:lavie:meteor-wmt:2014}, and BERTScore \cite{bert-score}. BERTScore uses pretrained contextual embeddings and cosine similarity to measure similarity between reference and hypothesis for text generation tasks. 

\subsection{Model Details}

The ASR models are trained using ESPNet \cite{watanabe2018espnet}. The speech encoder has 2x convolutional subsampling followed by 12 encoder layers, each with feed-forward dimension of 2048, and 8 attention heads. The transformer decoder has 6 layers, each with feed-forward dimension 512 and 4 attention heads. ASR models are trained with joint Connectionist Temporal Classification (CTC)-Attention \cite{kim2017joint} with the weight for CTC training set to 0.3. Adam optimizer is used with a peak learning rate of 0.002 in 25k training steps.
The videos are trimmed to 100s for the video-level speech tasks owing to compute constraints. Batches of 20,000 frames/GPU are constructed for video-level training. Specaugment  \cite{park2019-specaugment} is used during model training and fine-tuning. We use 40-dimension filterbank and 3-dimensional pitch features for training all models. 

The Huggingface transformers library  \cite{wolf-etal-2020-transformers} is used to fine-tune text-only models. \texttt{BART-large} and \texttt{BART-base}  \cite{lewis-etal-2020-bart} are used to finetune the ATS model in the cascade approach, and the decoder from the model is used for E2E training of the speech summarization model.

\begin{table}[t]
\caption{Word Error Rate (WER) (\%) for Test and Held Test sets of the 2000h How-to Corpus. Window Size of 20 is used for Restricted Self-Attention}
\centering
\resizebox{0.5\textwidth}{!}{%
\begin{tabular}{lcc}
\hline
\textbf{Encoder} & \textbf{Decoder}      & \textbf{Test WER (\%)}   \\ \hline
Transformer     & Transformer &     10.2        \\ 
Conformer & Transformer      & \textbf{9.1}       \\ 
+ Restricted Self-Attention & Transformer &   9.3            \\ \hline
\end{tabular}%
}
\label{tab:asr_wer}
\end{table}

\begin{table}[h]
\caption{Effect of Window Size and Dilation in Self-Attention of the Speech Encoder on E2E Summarization Model Training. W is the Window Size, and D is the dilation factor (Section \ref{sec:proposed} for details).}
\centering
\resizebox{0.45\textwidth}{!}{%
\begin{tabular}{ccccc}
\hline
\textbf{W} & \textbf{D}      & \textbf{ROUGE-L} & \textbf{METEOR} & \textbf{BERTScore}  \\ \hline
20 & \xmark & 52.0 & 26.5 & 90.5\\
40 & \xmark & \textbf{53.1} & \textbf{27.3} & \textbf{90.6}\\
60 & \xmark & 52.5 & 27.1  & 90.5\\
100 & 5    & 51.9 & 26.3 & 90.5\\ 
 \hline
\end{tabular}
}
\label{tab:window_size}
\end{table}

\begin{table*}[h]
\caption{Summarization Performance of Topline, Cascade and E2E Models using automatic (ROUGE and METEOR) and semantic evaluation metrics (BERTScore).}
\begin{center}
\resizebox{1.0\textwidth}{!}{%
\begin{tabular}{llcccccc}
\hline
& \textbf{Model}  & \textbf{Parameters} & \textbf{ROUGE-1} & \textbf{ROUGE-2} & \textbf{ROUGE-L} & \textbf{METEOR} & \textbf{BERTScore} \\ 
\hline
\parbox[t]{3mm}{\multirow{3}{*}{\rotatebox[origin=c]{90}{Topline}}} & Groundtruth Text & & & &\\
& + \texttt{BART-large} Summarization &   400M   &    \textbf{61.8}        &     42.8             &       55.5           & 30.0 & \textbf{91.0}             \\ 
& + \texttt{BART-base} Summarization  & 140M &     60.6            &       40.4           &        53.7          &  27.7   &    90.7     \\ 
\hline
\parbox[t]{3mm}{\multirow{5}{*}{\rotatebox[origin=c]{90}{Cascade}}} & Conformer ASR & 107M & & &  \\
& + \texttt{BART-large} Summarization &   400M &       59.2       &  38.8               &               52.3   &  27.8 & 90.6          \\ 
& + \texttt{BART-base} Summarization & 140M & 57.6                &        36.3        & 50.3                &   25.6 & 90.3           \\ 
& S2S- PredText2Summary\cite{palaskar-etal-2019-multimodal} & - & - & - & 46.1 & 22.9 & - \\
& Kaldi ASR + Concept2Summary \cite{palaskar21_interspeech} & - & - & - & 51.4 & \textbf{30.4} & - \\ 
\hline
\parbox[t]{3mm}{\multirow{2}{*}{\rotatebox[origin=c]{90}{E2E}}} & Conformer Encoder & & & & \\
& + Transformer Decoder & \textbf{104M} & 60.9 & \textbf{43.0} & \textbf{55.9} & 28.8 & \textbf{91.0} \\
\hline
\end{tabular}
}
\label{tab:summ_results}
\end{center}
\end{table*}

\section{Results and Discussion}
\label{sec:results_discussion}
\subsection{Speech Recognition}

The E2E Speech Summarization model is pre-trained for ASR. Table \ref{tab:asr_wer} shows the Word Error Rate (WER) for different encoder-decoder combinations. The use of a conformer \cite{conformer2020} model improves ASR results by over 1 \% absolute compared to the transformer. The use of restricted self-attention results in a slight decrease in performance.

\subsection{Speech Summarization}

Table \ref{tab:summ_results} highlights summarization results on three types of models: ground-truth text-based models (considered the topline scores), ASR-based Cascade models, and direct E2E models.
Cascade models use the best ASR model from Table \ref{tab:asr_wer}, i.e., a conformer encoder and transformer decoder. \texttt{BART-large} and \texttt{BART-base}  \cite{lewis-etal-2020-bart} are fine-tuned on ground-truth and ASR predicted text to establish the topline and cascade baselines. \texttt{BART-large} outperforms \texttt{BART-base} in ROUGE, METEOR, and BERT Scores among the topline and cascade models. Conformer ASR coupled with BART leads to strong cascade models that outperform previous works.

The best ASR model is finetuned on the summarization data to build the E2E model with conformer encoder and transformer decoder. The E2E model outperforms the best cascade model on all metrics with 4x fewer parameters, indicating that the end-to-end model is able to produce more fluent, semantically similar summaries. It is interesting to note that the E2E model performs nearly as well the best topline model, indicating that the task of speech summarization can be performed just as well without text or transcribed speech.



\subsection{Window Size and Dilation}

To understand the impact of context window size on summarization performance, we train models with different window sizes using a subset of the training data. This subset consists of about 65 \% of the full training data, and untrimmed videos ( with video length $<=$ 100s ). Then the model is evaluated on the full test set. Table \ref{tab:window_size} shows that summarization performance does depend on window size. A window size of $W=40$ seems to yield the best ROUGE-L scores, while a smaller window of $W=20$ yields a lower ROUGE-L score. From the first and last row, dilation reduces the computational complexity significantly while retaining similar performance.






\subsection{Concept Learning}


Table \ref{tab:concept_eval} evaluates the end-to-end concept learning model. Concepts being non-sequential text, we evaluate on Precision, Recall, and F1. The baseline is a cascade of two modules- ASR and   \textit{predicted} Text2Concept model, and achieves an F1 of 54.8 \cite{palaskar21_interspeech}. The proposed end-to-end Speech2Concept model achieves 57.2, a significant performance given the model sees no input text.

\begin{table}[t]
\caption{Evaluation of Baseline and Proposed Concept Learning Models using Recall, Precision and F-1 Score}
\centering
\resizebox{0.45\textwidth}{!}{%
\begin{tabular}{lccc}
\hline
\textbf{Model} & \textbf{Precision}      & \textbf{Recall} & \textbf{F-1}  \\ \hline
Predicted Text2Concept \cite{palaskar21_interspeech} & 52.5 & 57.3 & 54.8 \\
Speech2Concept & \textbf{62.3} & 55.8 & \textbf{58.8} \\
 \hline
\end{tabular}
}
\label{tab:concept_eval}
\end{table}

\section{Conclusion}
\label{sec:conclusion}
In this paper, we model speech summarization as a direct end-to-end task starting from input speech at a video-level and generating abstractive summaries as the output. We address the long speech input frames problem by applying restricted self-attention to help us achieve this task without running into severe memory and compute bottlenecks. Our approach at least outperforms a strong text-based summarization model, and at best, demonstrates strong performance compared to previous approaches to speech summarization (cascaded pipeline models). We also demonstrate the effects of various window size and dilations on summarization, concluding that larger window sizes are crucial for better models.  Using restricted self-attention and a Conformer based speech recognizer, we achieve a competitive result on speech recognition on the commonly used How2 dataset. Finally, we demonstrate the potential of such end-to-end modeling on a Speech2Concept task that could be useful for downstream summarization as well as other speech-based tasks that earlier represented speech by predicted text from an automatic speech recognizer.


\section{Acknowledgements}
This work used the Bridges system, which is supported by
NSF award number ACI-1445606, at the Pittsburgh Supercomputing Center (PSC).

\small
\bibliographystyle{IEEEbib}
\bibliography{template}

\begin{thebibliography}{10}

\bibitem{Hahn2002}
U.~Hahn and I.~Mani,
\newblock ``The challenges of automatic summarization,''
\newblock {\em Computer}, vol. 33, no. 11, pp. 29--36, 2000.

\bibitem{palaskar-etal-2019-multimodal}
Shruti Palaskar, Jind{\v{r}}ich Libovick{\'y}, Spandana Gella, and Florian
  Metze,
\newblock ``Multimodal abstractive summarization for how2 videos,''
\newblock in {\em Proceedings of the 57th Annual Meeting of the Association for
  Computational Linguistics}, Florence, Italy, 2019, pp. 6587--6596,
  Association for Computational Linguistics.

\bibitem{yu2021vision}
Tiezheng Yu, Wenliang Dai, Zihan Liu, and Pascale Fung,
\newblock ``Vision guided generative pre-trained language models for multimodal
  abstractive summarization,'' 2021.

\bibitem{zhu2020hierarchical}
Chenguang Zhu, Ruochen Xu, Michael Zeng, and Xuedong Huang,
\newblock ``A hierarchical network for abstractive meeting summarization with
  cross-domain pretraining,''
\newblock in {\em Findings of the Association for Computational Linguistics:
  EMNLP 2020}, Online, 2020, pp. 194--203, Association for Computational
  Linguistics.

\bibitem{rezazadegan2020automatic}
Dana Rezazadegan, Shlomo Berkovsky, Juan~C Quiroz, A~Baki Kocaballi, Ying Wang,
  Liliana Laranjo, and Enrico Coiera,
\newblock ``Automatic speech summarisation: A scoping review,''
\newblock {\em arXiv preprint arXiv:2008.11897}, 2020.

\bibitem{manakul2020abstractive}
Potsawee Manakul, Mark Gales, and Linlin Wang,
\newblock ``Abstractive spoken document summarization using hierarchical model
  with multi-stage attention diversity optimization,''
\newblock 2020.

\bibitem{palaskar21_interspeech}
Shruti Palaskar, Ruslan Salakhutdinov, Alan~W. Black, and Florian Metze,
\newblock ``{Multimodal Speech Summarization Through Semantic Concept
  Learning},''
\newblock in {\em Proc. Interspeech 2021}, 2021, pp. 791--795.

\bibitem{dai-etal-2019-transformer}
Zihang Dai, Zhilin Yang, Yiming Yang, Jaime Carbonell, Quoc Le, and Ruslan
  Salakhutdinov,
\newblock ``Transformer-{XL}: Attentive language models beyond a fixed-length
  context,''
\newblock in {\em Proceedings of the 57th Annual Meeting of the Association for
  Computational Linguistics}, Florence, Italy, 2019, pp. 2978--2988,
  Association for Computational Linguistics.

\bibitem{Rae2019CompressiveTransformer}
Jack~W. Rae, Anna Potapenko, Siddhant~M. Jayakumar, Chloe Hillier, and
  Timothy~P. Lillicrap,
\newblock ``Compressive transformers for long-range sequence modelling,''
\newblock in {\em 8th International Conference on Learning Representations,
  {ICLR} 2020, Addis Ababa, Ethiopia, April 26-30, 2020}. 2020, OpenReview.net.

\bibitem{kitaev2019reformer}
Nikita Kitaev, Lukasz Kaiser, and Anselm Levskaya,
\newblock ``Reformer: The efficient transformer,''
\newblock in {\em 8th International Conference on Learning Representations,
  {ICLR} 2020, Addis Ababa, Ethiopia, April 26-30, 2020}. 2020, OpenReview.net.

\bibitem{ainslie-etal-2020-etc}
Joshua Ainslie, Santiago Ontanon, Chris Alberti, Vaclav Cvicek, Zachary Fisher,
  Philip Pham, Anirudh Ravula, Sumit Sanghai, Qifan Wang, and Li~Yang,
\newblock ``{ETC}: Encoding long and structured inputs in transformers,''
\newblock in {\em Proceedings of the 2020 Conference on Empirical Methods in
  Natural Language Processing (EMNLP)}, Online, 2020, pp. 268--284, Association
  for Computational Linguistics.

\bibitem{wang2020linformer}
Sinong Wang, Belinda Li, Madian Khabsa, Han Fang, and Hao Ma,
\newblock ``Linformer: Self-attention with linear complexity,''
\newblock {\em arXiv preprint arXiv:2006.04768}, 2020.

\bibitem{Zaheer2020BigBird}
Manzil Zaheer, Guru Guruganesh, Kumar~Avinava Dubey, Joshua Ainslie, Chris
  Alberti, Santiago Ontanon, Philip Pham, Anirudh Ravula, Qifan Wang, Li~Yang,
  and Amr Ahmed,
\newblock ``Big bird: Transformers for longer sequences,''
\newblock in {\em Advances in Neural Information Processing Systems},
  H.~Larochelle, M.~Ranzato, R.~Hadsell, M.~F. Balcan, and H.~Lin, Eds. 2020,
  vol.~33, pp. 17283--17297, Curran Associates, Inc.

\bibitem{Beltagy2020Longformer}
Iz~Beltagy, Matthew~E. Peters, and Arman Cohan,
\newblock ``Longformer: The long-document transformer,''
\newblock {\em CoRR}, vol. abs/2004.05150, 2020.

\bibitem{Shafey2019}
Laurent~El Shafey, Hagen Soltau, and Izhak Shafran,
\newblock ``{Joint Speech Recognition and Speaker Diarization via Sequence
  Transduction},''
\newblock in {\em Proc. Interspeech 2019}, 2019, pp. 396--400.

\bibitem{HoriMHR20}
Takaaki Hori, Niko Moritz, Chiori Hori, and Jonathan~Le Roux,
\newblock ``Transformer-based long-context end-to-end speech recognition,''
\newblock in {\em Interspeech 2020, 21st Annual Conference of the International
  Speech Communication Association, Virtual Event, Shanghai, China, 25-29
  October 2020}, Helen Meng, Bo~Xu, and Thomas~Fang Zheng, Eds. 2020, pp.
  5011--5015, {ISCA}.

\bibitem{hori2021advanced}
Takaaki Hori, Niko Moritz, Chiori Hori, and Jonathan~Le Roux,
\newblock ``Advanced long-context end-to-end speech recognition using
  context-expanded transformers,'' 2021.

\bibitem{Masumura2021}
Ryo Masumura, Naoki Makishima, Mana Ihori, Akihiko Takashima, Tomohiro Tanaka,
  and Shota Orihashi,
\newblock ``Hierarchical transformer-based large-context end-to-end asr with
  large-context knowledge distillation,''
\newblock in {\em ICASSP 2021 - 2021 IEEE International Conference on
  Acoustics, Speech and Signal Processing (ICASSP)}, 2021, pp. 5879--5883.

\bibitem{lewis-etal-2020-bart}
Mike Lewis, Yinhan Liu, Naman Goyal, Marjan Ghazvininejad, Abdelrahman Mohamed,
  Omer Levy, Veselin Stoyanov, and Luke Zettlemoyer,
\newblock ``{BART}: Denoising sequence-to-sequence pre-training for natural
  language generation, translation, and comprehension,''
\newblock in {\em Proceedings of the 58th Annual Meeting of the Association for
  Computational Linguistics}, Online, 2020, pp. 7871--7880, Association for
  Computational Linguistics.

\bibitem{Moritz2021}
Niko Moritz, Takaaki Hori, and Jonathan Le~Roux,
\newblock ``Capturing multi-resolution context by dilated self-attention,''
\newblock in {\em ICASSP 2021 - 2021 IEEE International Conference on
  Acoustics, Speech and Signal Processing (ICASSP)}, 2021, pp. 5869--5873.

\bibitem{sanabria2018how2}
Ramon Sanabria, Ozan Caglayan, Shruti Palaskar, Desmond Elliott, Lo{\"\i}c
  Barrault, Lucia Specia, and Florian Metze,
\newblock ``How2: a large-scale dataset for multimodal language
  understanding,''
\newblock {\em arXiv preprint arXiv:1811.00347}, 2018.

\bibitem{lin-2004-rouge}
Chin-Yew Lin,
\newblock ``{ROUGE}: A package for automatic evaluation of summaries,''
\newblock in {\em Text Summarization Branches Out}, Barcelona, Spain, 2004, pp.
  74--81, Association for Computational Linguistics.

\bibitem{denkowski:lavie:meteor-wmt:2014}
Michael Denkowski and Alon Lavie,
\newblock ``Meteor universal: Language specific translation evaluation for any
  target language,''
\newblock in {\em Proceedings of the Ninth Workshop on Statistical Machine
  Translation}, Baltimore, Maryland, USA, 2014, pp. 376--380, Association for
  Computational Linguistics.

\bibitem{bert-score}
Tianyi Zhang, Varsha Kishore, Felix Wu, Kilian~Q. Weinberger, and Yoav Artzi,
\newblock ``Bertscore: Evaluating text generation with {BERT},''
\newblock in {\em 8th International Conference on Learning Representations,
  {ICLR} 2020, Addis Ababa, Ethiopia, April 26-30, 2020}. 2020, OpenReview.net.

\bibitem{watanabe2018espnet}
Shinji Watanabe, Takaaki Hori, Shigeki Karita, Tomoki Hayashi, Jiro Nishitoba,
  Yuya Unno, Nelson {Enrique Yalta Soplin}, Jahn Heymann, Matthew Wiesner,
  Nanxin Chen, Adithya Renduchintala, and Tsubasa Ochiai,
\newblock ``{ESPnet}: End-to-end speech processing toolkit,''
\newblock in {\em Proceedings of Interspeech}, 2018, pp. 2207--2211.

\bibitem{kim2017joint}
Suyoun Kim, Takaaki Hori, and Shinji Watanabe,
\newblock ``Joint ctc-attention based end-to-end speech recognition using
  multi-task learning,''
\newblock in {\em 2017 {IEEE} International Conference on Acoustics, Speech and
  Signal Processing, {ICASSP} 2017, New Orleans, LA, USA, March 5-9, 2017}.
  2017, pp. 4835--4839, {IEEE}.

\bibitem{park2019-specaugment}
Daniel~S. Park, William Chan, Yu~Zhang, Chung-Cheng Chiu, Barret Zoph,
  Ekin~Dogus Cubuk, and Quoc~V. Le,
\newblock ``Specaugment: A simple augmentation method for automatic speech
  recognition,''
\newblock in {\em INTERSPEECH}, 2019.

\bibitem{wolf-etal-2020-transformers}
Thomas Wolf, Lysandre Debut, Victor Sanh, Julien Chaumond, Clement Delangue,
  Anthony Moi, Pierric Cistac, Tim Rault, Remi Louf, Morgan Funtowicz, Joe
  Davison, Sam Shleifer, Patrick von Platen, Clara Ma, Yacine Jernite, Julien
  Plu, Canwen Xu, Teven Le~Scao, Sylvain Gugger, Mariama Drame, Quentin Lhoest,
  and Alexander Rush,
\newblock ``Transformers: State-of-the-art natural language processing,''
\newblock in {\em Proceedings of the 2020 Conference on Empirical Methods in
  Natural Language Processing: System Demonstrations}, Online, 2020, pp.
  38--45, Association for Computational Linguistics.

\bibitem{conformer2020}
Anmol Gulati, Chung-Cheng Chiu, James Qin, Jiahui Yu, Niki Parmar, Ruoming
  Pang, Shibo Wang, Wei Han, Yonghui Wu, Yu~Zhang, and Zhengdong Zhang, Eds.,
\newblock {\em Conformer: Convolution-augmented Transformer for Speech
  Recognition}, 2020.

\end{thebibliography}

\end{document}